\documentclass[12pt, letterpaper]{article}
\usepackage[utf8]{inputenc}
\usepackage{lipsum}
\usepackage{hyperref}
\usepackage{amssymb}
\usepackage{amsmath}
\usepackage{graphicx}
\usepackage{algorithm}
\usepackage{algpseudocode}

\title{Addressing computational challenges in physical system simulations with machine learning}
\author{
    Sabber Ahamed$^{1a,*}$, Md Mesbah Uddin$^{1b,*}$ \\
    \texttt{$^{1a,*}$sabbers@gmail.com}, \texttt{$^{1b,*}$Mesbahuddin1991@gmail.com} \\
}

\begin{document}

\maketitle
\let\thefootnote\relax\footnote{{$^*$}These authors contributed equally to this work.}

\begin{abstract}
    \noindent In this paper, we present a machine learning-based data generator framework, tailored to aid researchers who utilize simulations to examine various physical systems or processes. High computational costs and the resulting limited data often pose significant challenges to gaining insights into these systems or processes. Our approach involves a two-step process: initially, we train a supervised predictive model using a limited simulated dataset to predict simulation outcome. Subsequently, a reinforcement learning agent is trained to generate accurate, simulation-like data by leveraging the supervised model. With this framework, researchers can generate more accurate data and know the outcomes without running high computational simulations. Which enables them to explore the parameter space more efficiently and gain deeper insights into physical systems or processes. In this paper, we demonstrate the effectiveness of the proposed framework by applying it to two case studies
    one focusing on earthquake rupture physics and the other on new material development.
\end{abstract}

\maketitle

\section{Introduction}

Understanding physical systems or processes often involves the use of computational simulations, a method that can be both time-consuming and computationally expensive. For instance, simulating the weather and climate dynamics~\cite{washington2005introduction}, modeling complex biological systems like protein folding~\cite{shaw2010atomic}, or even understaing earthquake rupture process~\cite{harris2018suite} or seismic activities ~\cite{graves2011cybershake, shaw2018physics} requires a significant computational demands.These constraints can severely limit the amount of data available for research and impact our ability to gain deep insights into these systems or processes.

In response to these challenges, we propose a novel machine learning-based generator framework designed to enhance the efficiency of data generation in studies that rely on simulations. This framework operates in a two-step approach. Initially, a supervised predictive model is trained using a small-scale simulated dataset with varying input parameters. This predictive model, functioning as a surrogate for the original physical simulations, is then used to train a reinforcement learning (RL) agent \cite{sutton2018reinforcement}. The RL agent, guided by the feedback from the predictive model, learns to generate more accurate, simulation-like data.

The key advantage of this framework is that it facilitates the generation of larger quantities of data without the need for further high computational simulations. This allows researchers to explore the parameter space more efficiently, thereby gaining deeper insights into the physical systems or processes under study. Furthermore, by employing a reinforcement learning approach, our framework can continually adapt and improve over time, offering the potential for even greater accuracy and efficiency in future data generation.

In this paper, we demonstrate the effectiveness of the proposed framework by applying it to two case studies in material science and geodynamics, thereby highlighting its versatility and broad applicability across various domains of physical sciences.

\section{Generator Framework}

The proposed machine learning framework consists of two main components: (1) a supervised predictive model, (2) a reinforcement learning agent. The framework is consists of these interdependent components that work together to generate more simulations like data and predict the outcomes of simulations.

\subsection{Overview}

\begin{figure}[!ht]
    \begin{center}
        \includegraphics[scale=0.60]{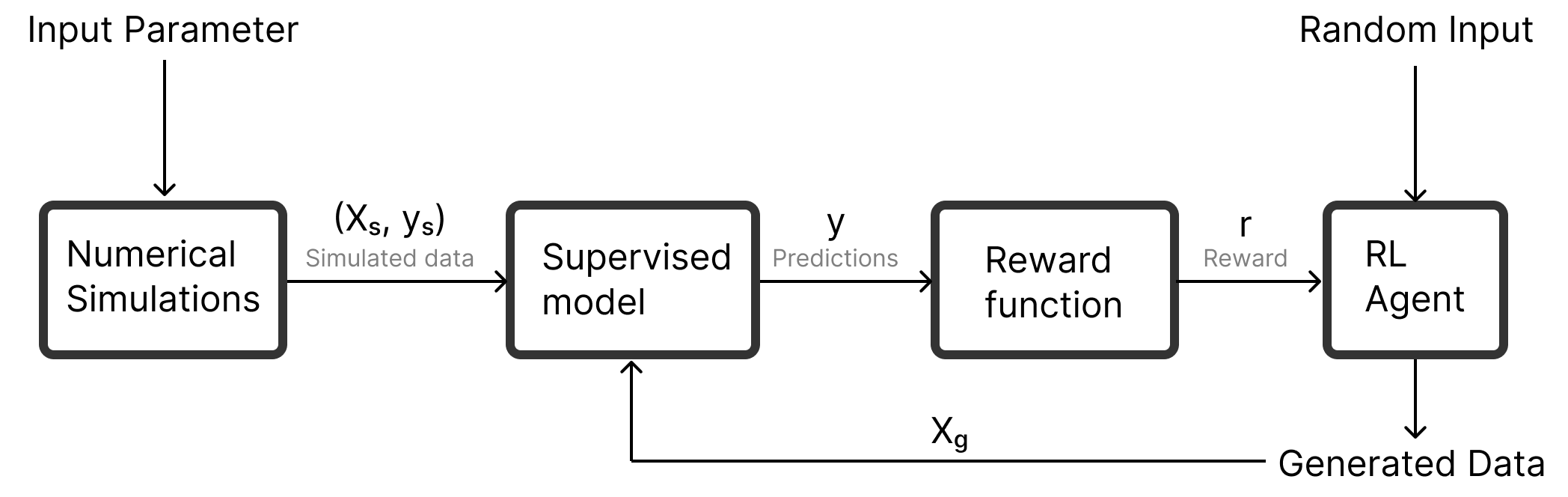}
    \end{center}
    \caption{The schematic diagram shows how the framework works}
    \label{fig:framework-diagram}
\end{figure}

First, we generate some simulated dataset with varying input parameters ($X_s$). We then train a supervised predictive model, $f(X_s)$ to predict the simulation outcomes ($y_s$). In the second step, we train a reinforcement learning agent with policy $\pi_{\theta}(a|s)$ using the supervised model to produce more accurate, simulation-like data. Where $a$ represents the action (generated data $X_g$) and $s$ is the random input parameter ($s\sim \mathcal{N}$) for the agent. $\theta$ in the policy $\pi_{\theta}(a|s)$ represents the parameters of the policy.

The agent's objective is to learn a policy that maximizes the outcome predicted by the supervised model. The reinforcement learning agent achieves this by adjusting the policy parameters $\theta$ to increase the probability of actions that lead to higher predicted outcomes~\cite{sutton2018reinforcement}. In the following subsections, we describe each component, along with the underlying models and methods.

\subsection{Supervised Model: Predicting the Outcome of Simulations}

Our framework's core is the supervised predictive model that predicts simulation outcomes ($y_s$) based on given input parameters ($X_s$). It functions as a map, approximating the genuine relationship between the inputs and outcomes, and can be applied to both classification and regression tasks.

The model's performance significantly influences the subsequent reinforcement learning agent's effectiveness. A well-performing model is vital for generating simulation-like data as the agent bases its data generation process on the model's predictions. If the model is inaccurate, it could misdirect the agent, leading to less realistic data. Therefore, a robust and accurate model is crucial in our framework, impacting the quality of the generated data and the framework's overall efficiency.

\subsection{RL Agent: Generate simulation like data}

In our framework, the reinforcement learning agent learns how to generate data from random inputs (Figure-\ref{fig:framework-diagram}). This process is not arbitrary; it's guided by the supervised model, which ensures the generated parameters are contextually relevant. Through iterative learning and exploration, the agent develops an ability to manipulate these random inputs in a way that can maximize or minimize the outcome as predicted by the supervised model. This learning process enables the agent to produce data that closely aligns with the complex dynamics of the physical system or process under study.

\subsubsection{Reward Function}

The reward function, typically denoted as $R(s,a)$, is a vital component of the reinforcement learning (RL) process~\cite{abbeel2004apprenticeship, brunskill2013sample, ng2000algorithms}. Here, $s$ represents the current state and $a$ represents the action taken. In our context, the reward function is designed to incentivize the RL agent to generate high-quality input parameter combinations that lead to desirable simulation outcomes.

The design of the reward function can vary depending on the specific objectives of the simulation~\cite{abbeel2004apprenticeship}. For instance, if the aim is to minimize the outcome predicted by the supervised model, denoted as $f(X_s)$, the reward function can be structured to give higher rewards for actions resulting in lower predicted outcomes. Conversely, if the aim is to maximize the outcome, the reward function can be designed to provide greater rewards for actions that result in higher predicted outcomes.

This flexibility in defining the reward function allows our RL agent to adapt to various objectives, making our framework versatile and capable of generating high-quality data for a broad range of tasks~\cite{sutton2018reinforcement, mnih2015human, lapan2018deep, karpathy2016deep}.

\subsubsection{Policy Learning for Data Generation}

In our proposed framework, the reinforcement learning (RL) agent learns a policy, denoted as $\pi_{\theta}(a|s)$, that determines the likelihood of choosing a particular action $a$ (or set of input parameters), given the current state $s$. Here, $\theta$ denotes the parameters of the policy. The goal of the RL agent is to learn an optimal policy, denoted as $\pi^*_{\theta}(a|s)$, that maximizes the expected cumulative reward, thereby leading to the generation of accurate, simulation-like data.

The learning process is guided by the Bellman equation for policy iteration~\cite{sutton2018reinforcement}:

\begin{equation}
    V^\pi(s) = \sum_{a} \pi_{\theta}(a|s) \left[R(s,a) + \gamma \sum_{s'} P(s'|s,a) V^\pi(s')\right]
\end{equation}

In this equation, $V^\pi(s)$ represents the value of state $s$ under policy $\pi$, $R(s,a)$ is the reward obtained by performing action $a$ in state $s$, $P(s'|s,a)$ denotes the probability of transitioning to state $s'$ from state $s$ after taking action $a$, and $\gamma$ is the discount factor, which determines the present value of future rewards.

Through iterative updates of the value function and the policy, guided by the Bellman equation, the RL agent gradually learns the optimal policy. This learned policy can generate the best actions for each state, leading to the generation of high-quality, simulation-like data. For a more detailed discussion of the reinforcement learning process, please refer to \cite{sutton2018reinforcement}.

\section{Case Studies}
In this section, we will demonstrate the application and effectiveness of our framework through a two case studies. The first case study focuses on the field of earthquake rupture physics, a complex and data-scarce area that significantly benefits from our approach. The second case study focuses on the field of material science, where our framework can be used to optimize processing conditions for developing high-performance materials.

\subsection{Earthquake Rupture Physics}
Simulating dynamic earthquake rupture propagation poses significant challenges due to uncertainties in fault slip physics, stress conditions, and frictional properties. Numerical simulations, while essential in understanding rupture physics, are highly dependent on initial parameters, which are difficult to optimize given the vast parameter space. As a result,researchers often resort to simplifying assumptions or trial-and-error methods to generate simulations that can overlook complexities and are computationally expensive~\cite{douilly20153d, ripperger2008variability, peyrat2001dynamic, ahamed2019estimating}

\subsubsection{Rupture simulations and data processing}
\begin{figure}[ht]
    \begin{center}
        \includegraphics[scale=0.6]{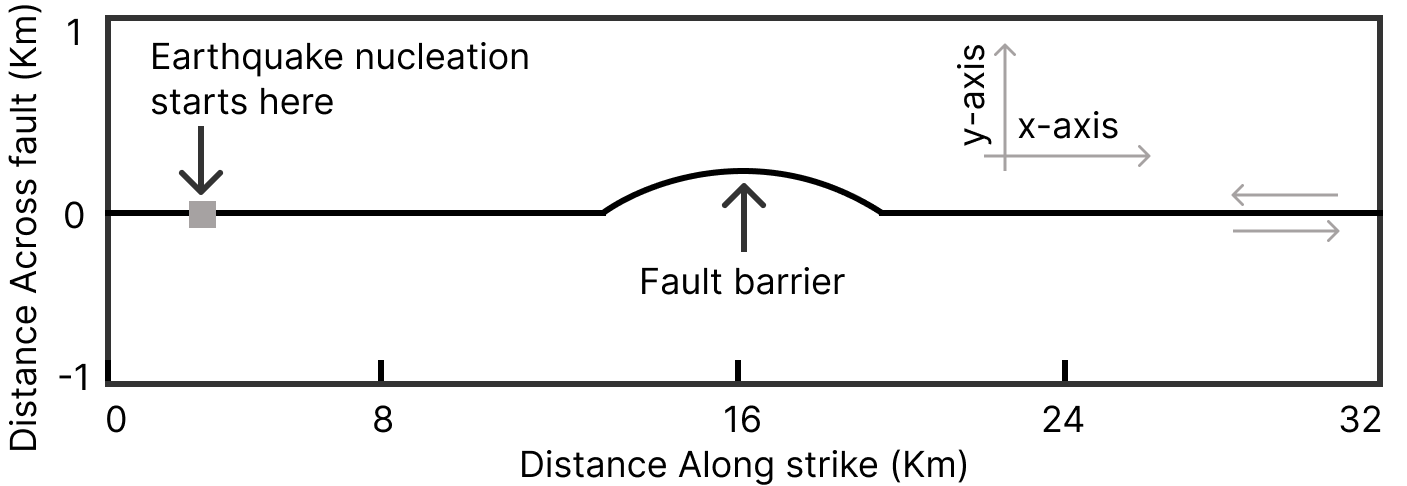}
    \end{center}
    \caption{A zoomed view of the two-dimensional fault geometry (not to the scale). The domain is 32 km long along the strike of the fault and 24 kilometers wide across the fault. The rupture starts to nucleate 10 km to the left of the barrier and propagates from the hypocenter towards the barrier}
    \label{fig:rupture_domain}
\end{figure}

In this example, we used 2000 simulated earthquake ruptures that were created by~\cite{ahamed2021application} to predict if an earthquake can break through a fault with geometric heterogeneity (Figure.~\ref{fig:rupture_domain}). The domain is 32 km long and 24 km wide. An open-source \texttt{C++} and \texttt{python} based library $\texttt{fdfault}$~\cite{fdfaultEricGithub} was used to generate the ruptures.

In each simulation, eight parameters were varied: x and y components of normal stress ($\sigma_{xx}$ and $\sigma_{yy}$), shear stress ($\sigma_{xy}$), dynamic friction coefficient, friction drop ($\mu_{s} - \mu_{d}$), critical slip distance ($d_c$), and width and height of the geometric heterogeneity at the center.

\subsubsection{Supervised Model to Predict Earthquake Rupture}

We used \texttt{LightGBM}~\cite{ke2017lightgbm}, a gradient boosting decision tree-based algorithm, to train a supervised machine learning model with the objective of predicting earthquake rupture outcomes. This training was conducted on a dataset of 1600 simulation instances, with an additional 400 simulations used for model validation. The target variable for prediction was the binary rupture outcome, coded as '1' when the rupture successfully breaks through the barrier, and '0' when it does not.

To enhance the model's predictive capabilities, we created additional features. These included the ratio of width to height, the difference in normal stresses ($\sigma_{xx} - \sigma_{yy}$), and the friction product ($\mu_{d} - sdrop$), among others. These features were derived from the original parameters to capture more complex relationships in the data.

The model was then trained using the expanded feature set and evaluated on the test data. The model performed quite well, achieving a ROC-AUC score of 0.8991 and a macro F1 score of 0.8266. The confusion matrix showed a good balance between sensitivity and specificity, with 100 true positives, 239 true negatives, 33 false positives, and 28 false negatives.

\subsubsection{Reinforcement Learning to Generate Rupture Parameters}

We used \texttt{Stable-baseline-3}~\cite{stable-baselines3}, a reinforcement learning library, to train an RL agent. The agent's goal was to create rupture parameters that would produce a rupture outcome mirroring that of the training data. We trained the RL agent using the Proximal Policy Optimization (PPO) algorithm\cite{schulman2017proximal}, a commonly preferred method due to its effective balance between sample complexity and computational demand.

The reward function is crucial in this setup. We used supervised model to guide the RL agent by providing feedback on the quality of the generated parameters. The generated parameters are used to predict an earthquake rupture outcome using the supervised model. If the predicted outcome is out of the valid range or if the generated parameters are physically implausible (for example, if height or width is negative), a negative reward is given. Otherwise, a positive reward is granted, encouraging the RL agent to generate similar parameter combinations in the future. This mechanism ensures the RL agent learns to produce plausible and high-quality data over time, enhancing the effectiveness of our earthquake rupture prediction framework.

\subsubsection{Results and Insights}
\begin{figure}[ht]
    \begin{center}
        \includegraphics[scale=0.5]{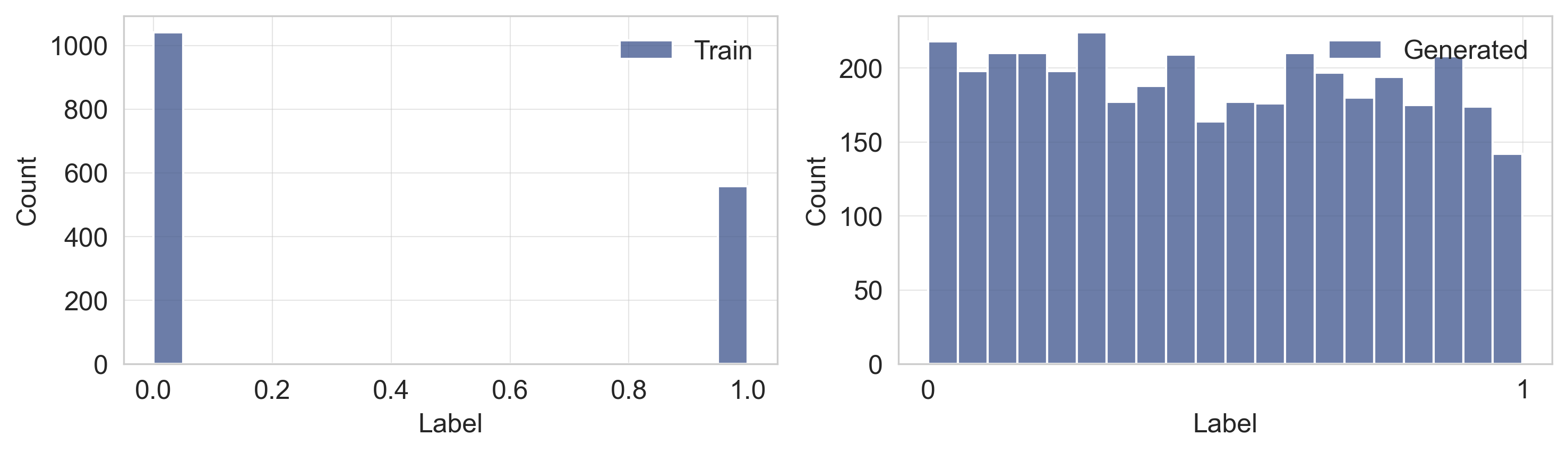}
    \end{center}
    \caption{Histogram plot representing rupture outcomes. (a) displays the distribution of outcomes in the original training data, which are confined to binary labels of 0 and 1, signifying specific classes of rupture. (b) illustrates the outcomes generated by reinforcement learning agent, which are spread continuously over a range from 0 to 1.}
    \label{fig:label_distribution}
\end{figure}

We used our reinforcement learning (RL) agent to generate 5000 data points, from which we filtered those that fall within the prediction range of the supervised model. As illustrated in Figure-\ref{fig:label_distribution}, the data produced by the RL agent spans a broader and more nuanced spectrum of values between 0 and 1, in contrast to the training data which is comprising of only 0s and 1s. This wider range of outcomes generated by the RL agent provides a more detailed insight into the complex rupture process, enhancing the predictive model's performance by offering varied and comprehensive training data. Thus, the RL agent's role in data generation is pivotal to improving the overall understanding and prediction of earthquake ruptures.

\begin{figure}[ht]
    \begin{center}
        \includegraphics[scale=0.45]{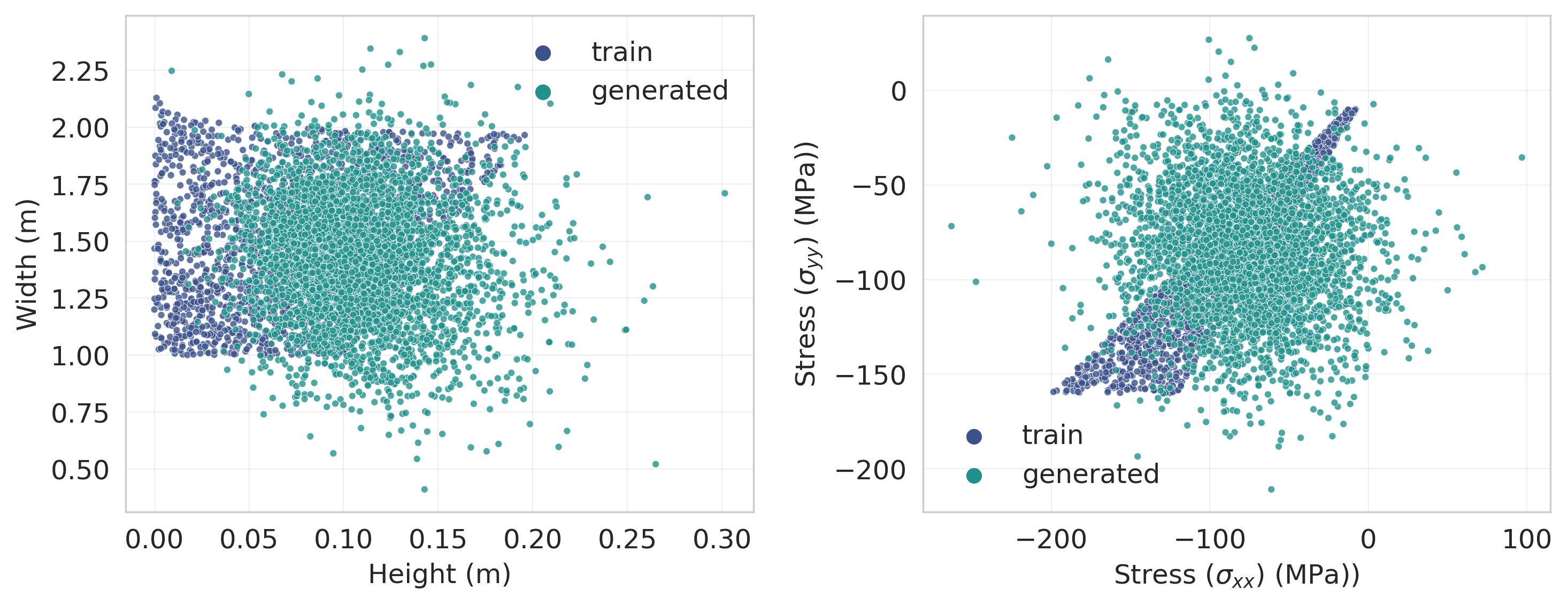}
    \end{center}
    \caption{2D scatter distribution plot of training and generated data, showing (a) Height and Width, and (b) Normal Stresses. The generated data exhibits a normal distribution due to the characteristics of our reinforcement learning (RL) environment. In the RL environment, the state is initialized and updated with a random draw from a standard normal distribution.}
    \label{fig:height_width_stress_scatter}
\end{figure}

Figure-\ref{fig:height_width_stress_scatter} presents a 2D scatter plot, demonstrating the correlation between Height and Width, as well as Normal Stresses within both the training and generated datasets. The generated data, a product of the RL agent's environment, displays a normal distribution. This is attributed to the state initialization and action space definition in the RL environment. The state, regularly refreshed with values from a standard normal distribution, coupled with a standard normal transformation applied to the data, ensures the normal distribution of the generated outcome. This approach allows the RL agent to explore a wide array of possible states, thereby producing a diverse set of generated data, allowing the RL agent to explore a broad spectrum of possible states.

\begin{figure}[!ht]
    \begin{center}
        \includegraphics[scale=0.50]{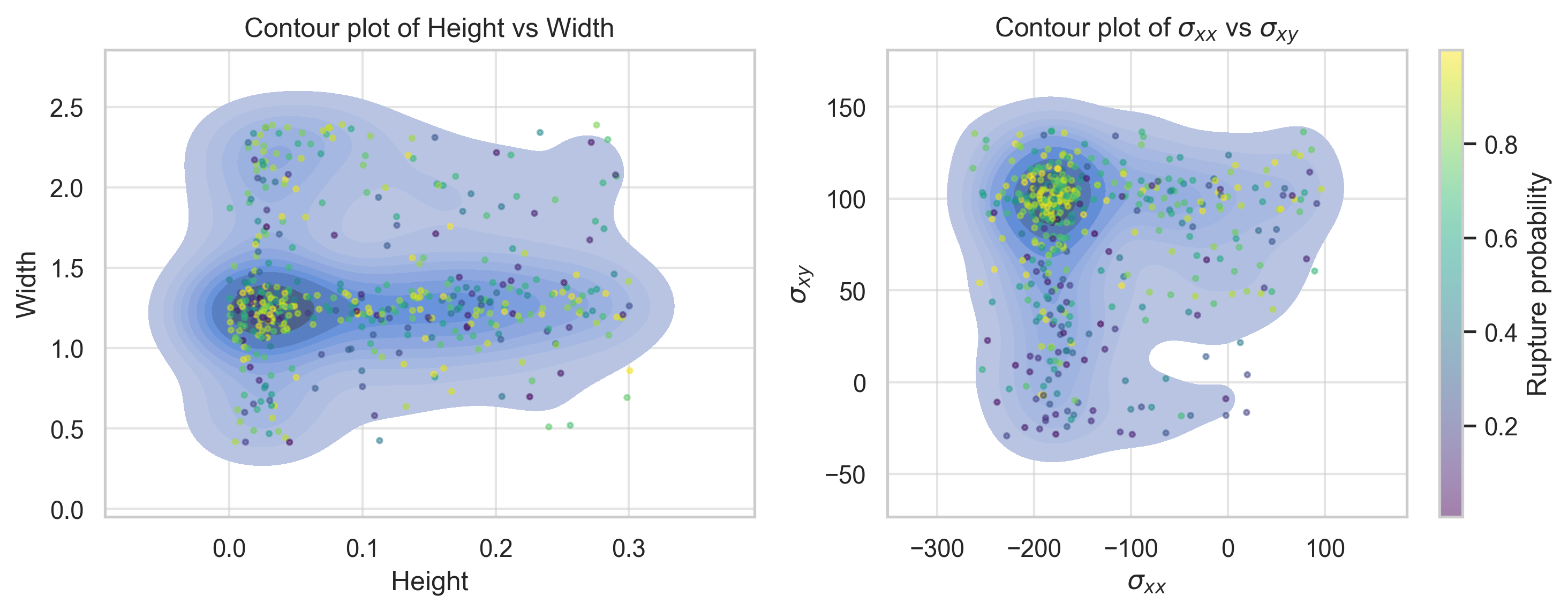}
    \end{center}
    \caption{Plot shows the contour of the rupture outcome. The dots with color represents the probability(a) height and width of the fault barrier (b) normal and shear stress of the fault}
    \label{fig:contour}
\end{figure}

Using the diverse generated data and parameter space provided by the RL agent, we were able to establish a range of plausible values for each parameter - these ranges being defined by the minimum and maximum values. Utilizing these information we used bayesian based optimization to understand the individual impacts of various parameters on earthquake rupture. During the optimization process, each trial involved selecting a combination of parameters within these defined ranges. The objective was to find the combination that maximized the rupture outcome. By conducting this optimization over 1000 trials, we systematically explored the parameter space to identify the combinations that lead to the highest rupture outcomes.

The results, illustrated by the contour plots (Figure-\ref{fig:contour}), underscored the sensitivity of the rupture outcome to both geometric heterogeneities and stress conditions. Notably, the rupture outcome demonstrated high dependence on the height and width of the geometric heterogeneity and a stronger sensitivity to normal stress compared to shear stress.Through this optimization method, which leverages the diverse data generated by the RL agent, we obtained a comprehensive understanding of the intricate dynamics within the rupture process.

\subsection{Material Science}
Material design involves searching for the best solutions by exploring the design space using
material composition and hierarchical structure. Molecular dynamics (MD) simulations have been
employed to understand mechanical properties and design materials at the atomic and molecular level~\cite{uddin2023finite, uddin2020studying, mesbah2023study}, but
their computational and time constraints limit their application to a few nanoscale simulations, failing to
provide a comprehensive understanding. This approach also overlooks the mechanical behavior of
materials at larger scales which is relevant to their applications. To overcome this limitation, machine
learning models can generate new datasets, offering a more thorough comprehension of material behavior
across various size scales.

\subsubsection{Simulations and data processing}

\begin{figure}[!ht]
    \begin{center}
        \includegraphics[scale=0.30]{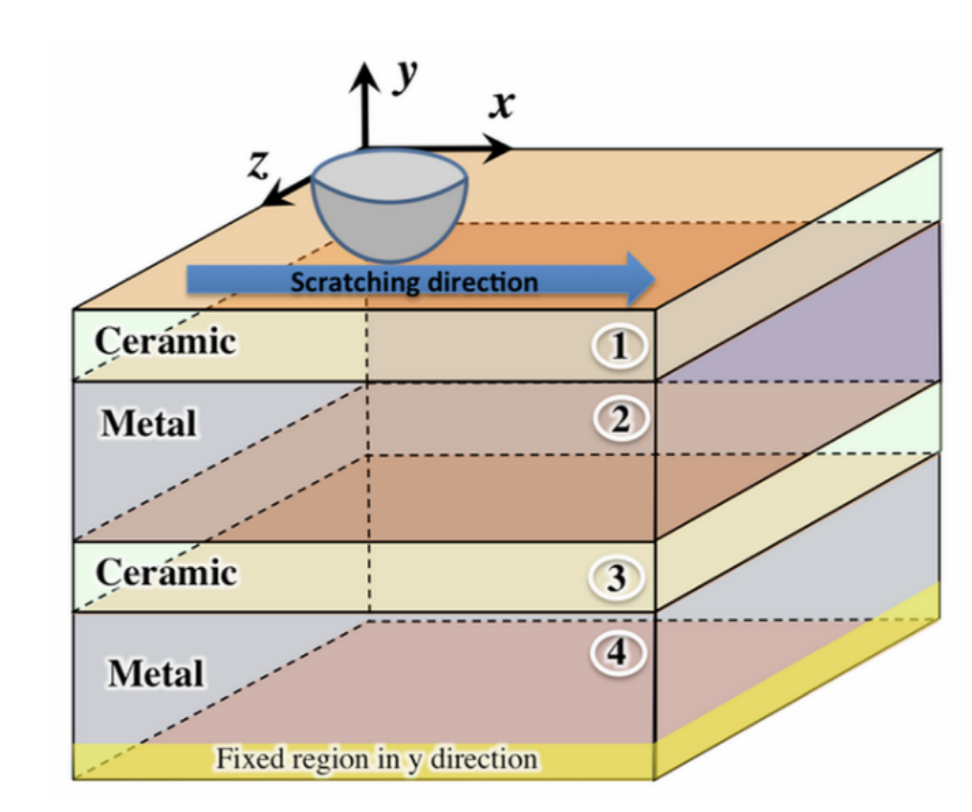}
    \end{center}
    \caption{Schematic diagram illustrating a simulation cell used for nanoindentation and scratching. The layers are numbered 1-4 from top to bottom. Only the thickness of layer 2 was varied.}
    \label{fig:material_design_domian}
\end{figure}

In this example, we used 18 models with different geometrical and loading parameters to predict
the scratching load, normal load, and friction coefficient. Multilayer samples were used, comprising
alternating layers of ceramic and metal (Figure-\ref{fig:material_design_domian}). The bottom metal layer had a fixed thickness of 6 nm
and acted as an elastic foundation for the layers above, mimicking the behavior of ceramic/metal
nanolaminates. The width of each multilayer was chosen to minimize strains and boundary effects, and
periodic boundary conditions were applied to the side faces. The thickness of the metallic and ceramic
layers was varied to investigate their impact on the mechanical and tribological properties of the samples.
Nano-indentation was performed using a rigid spherical nano-indentation with penetration depths ranging
from 3 nm to 7 nm, ensuring sufficient penetration of the metallic layers. For a penetration depth of 7 nm,
the minimum indenter radius was set to 10 nm (as a note, the indenter radius varied from 5 nm to 40 nm
depending on the penetration depth). The indenter speed was set to 100 m/s for nano-indentation and 250
m/s for nano-scratching, with a scratching length of 20 nm.

\subsubsection{Supervised Learning to Generate Material Data}
We also used \texttt{LightGBM}~\cite{ke2017lightgbm} to develop a supervised learning model to predict the frictional coefficient along the scratching distance.
The model was trained on 23 of these models, validated on 5, and tested on the remaining 7 models. To create a unique model, we combined the model layers with indenter radius and penetration depth. Then for each model, we collected frictional coefficient from different scratching data points.

\begin{figure}[!ht]
    \begin{center}
        \includegraphics[scale=0.30]{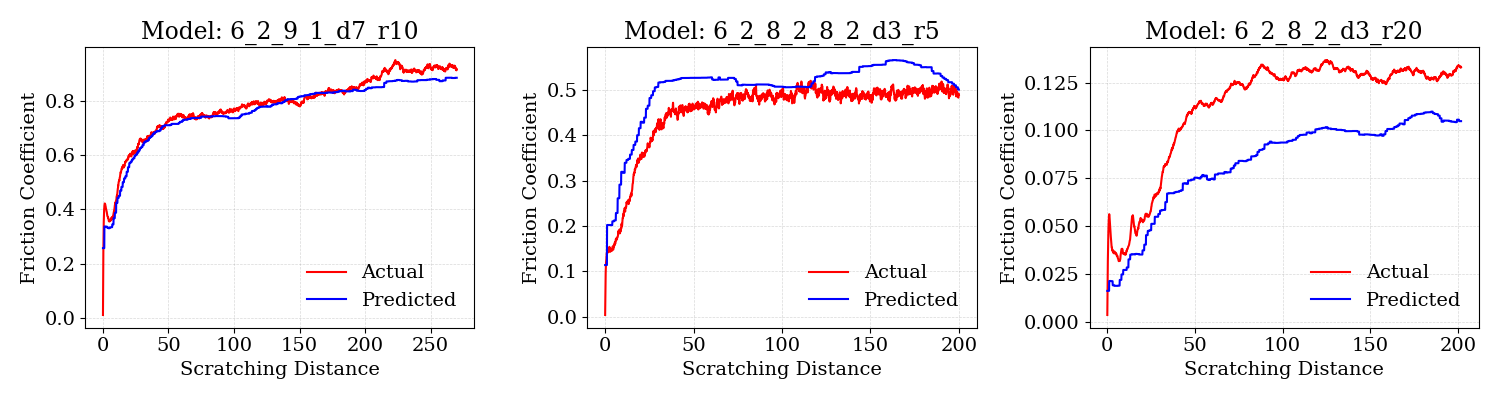}
    \end{center}
    \caption{Predicted frictional coefficient along the scratching distance for the test data on different models test simulations. Each unique model name, like '6\_2\_9\_1\_d7\_r10', encodes key properties: '6\_2\_9\_1' corresponds to the thickness of each material layer, 'd7' specifies a 7 nm indenter depth, and 'r10' represents a 10 nm indenter radius.}

    \label{fig:material_fric_pred}
\end{figure}

The performance of the supervised learning model was robust, with an R-squared value of 0.9581, a Mean Squared Error of 0.0030, Root Mean Squared Error of 0.0550 and Mean Absolute Error of 0.0455 on the test data.

Figure-\ref{fig:material_fric_pred} displays the predicted frictional coefficient over the course of scratching for various test models. Despite the limited size of the training dataset, the model appears to predict well across the test simulations. However, the figure also indicates a difficulty in accurately predicting models with larger indenter radius, as exemplified by the model where intender raius is 20.

\section{Limitations}

Our approach, while innovative, is not without limitations. The quality of the generated data and subsequent insights are tied to the quality of the training data provided to the RL agent. Any biases or gaps in this data could impact the efficacy of the RL agent's learning. Similarly, the supervised learning model's effectiveness is reliant on the richness and diversity of its training data. The current reward function design, although functional, may oversimplify the problem and limit the RL agent's ability to learn complex relationships or adapt for nuanced goals. Additionally, despite the RL agent's broad exploration capabilities, it is still constrained by the defined action and state spaces. Future work should aim to address these limitations, thereby enhancing the robustness and versatility of this model.

\section{Conclusion and Future Work}

The present work has demonstrated the innovative use of reinforcement learning, in generating a diverse parameter space. The designed custom environment and the reward system has proven to be effective in generating data that both supplements and extends the available training data, providing a broader perspective to the problem at hand.

Furthermore, the methodologies developed in this study have the potential to be applied across a range of other domains. The use of reinforcement learning and optimization techniques can be instrumental in exploring and understanding a plethora of complex systems. This work, thus, lays a solid foundation for harnessing the power of reinforcement learning and optimization in a broad spectrum of applications, opening up numerous exciting avenues for future research.

\bibliographystyle{unsrt}

\bibliography{ml_freamework_earthquake}

\end{document}